\documentclass[10pt,twocolumn,letterpaper]{article}

\usepackage[pagenumbers]{cvpr} 

\usepackage{graphicx}
\usepackage{amsmath}
\usepackage{amssymb}
\usepackage{booktabs}
\usepackage{amsfonts}
\usepackage{multirow}
\usepackage{comment}
\usepackage{balance}
\usepackage[accsupp]{axessibility}

\usepackage[pagebackref,breaklinks,colorlinks]{hyperref}

\usepackage[capitalize]{cleveref}
\crefname{section}{Sec.}{Secs.}
\Crefname{section}{Section}{Sections}
\Crefname{table}{Table}{Tables}
\crefname{table}{Tab.}{Tabs.}

\def\cvprPaperID{*****} 
\def\confName{CVPR}
\def\confYear{2022}

\begin{document}

\title{On Improving Cross-dataset Generalization of Deepfake Detectors}

\author{Aakash Varma Nadimpalli and Ajita Rattani\\
School of Computing \\
Wichita State University, Wichita, USA\\
{\tt\small axnadimpalli@shockers.wichita.edu,~ajita.rattani@wichita.edu}
}
\maketitle

\begin{abstract}
Facial manipulation by deep fake has caused major security risks and raised severe societal concerns. As a countermeasure, a number of deep fake detection methods have been proposed recently. Most of them model deep fake detection as a binary classification problem using a backbone convolutional neural network (CNN) architecture pretrained for the task. These CNN-based methods have demonstrated very high efficacy in deep fake detection with the Area under the Curve (AUC) as high as $0.99$. However, the performance of these methods degrades significantly when evaluated across datasets. In this paper, we formulate deep fake detection as a hybrid combination of supervised and reinforcement learning~(RL) to improve its cross-dataset generalization performance. The proposed method chooses the top-$k$ augmentations for each test sample by an RL agent in an image-specific manner. The classification scores, obtained using CNN, of all the augmentations of each test image are averaged together for final real or fake classification. Through extensive experimental validation, we demonstrate the superiority of our method over existing published research in cross-dataset generalization of deep fake detectors, thus obtaining state-of-the-art performance.
\end{abstract}

\section{Introduction}
\label{sec:intro}
Synthesized media, called deep fakes~\cite{citron, Tolosana2020deepfakesAB, Nguyen2019DeepLF}, containing facial information generated by digital manipulation techniques, have become a major political and societal threat~\cite{cellan-jones_2019}.  This term ``deep fake" signifies deep adversarial models that generate fake content by \emph{swapping} a person's face with the face of another person using deep fake generation techniques such as FaceSwap and FaceShifter~\cite{Li2019FaceShifterTH},~\cite{8237659}. The use of deep fakes to commit fraud, falsify evidence~\cite{Hwang2020}, manipulate public debates, and destabilize political processes has raised a top security concern. 




\begin{figure*}[htbp]
\centerline{\includegraphics[width=0.75\textwidth]{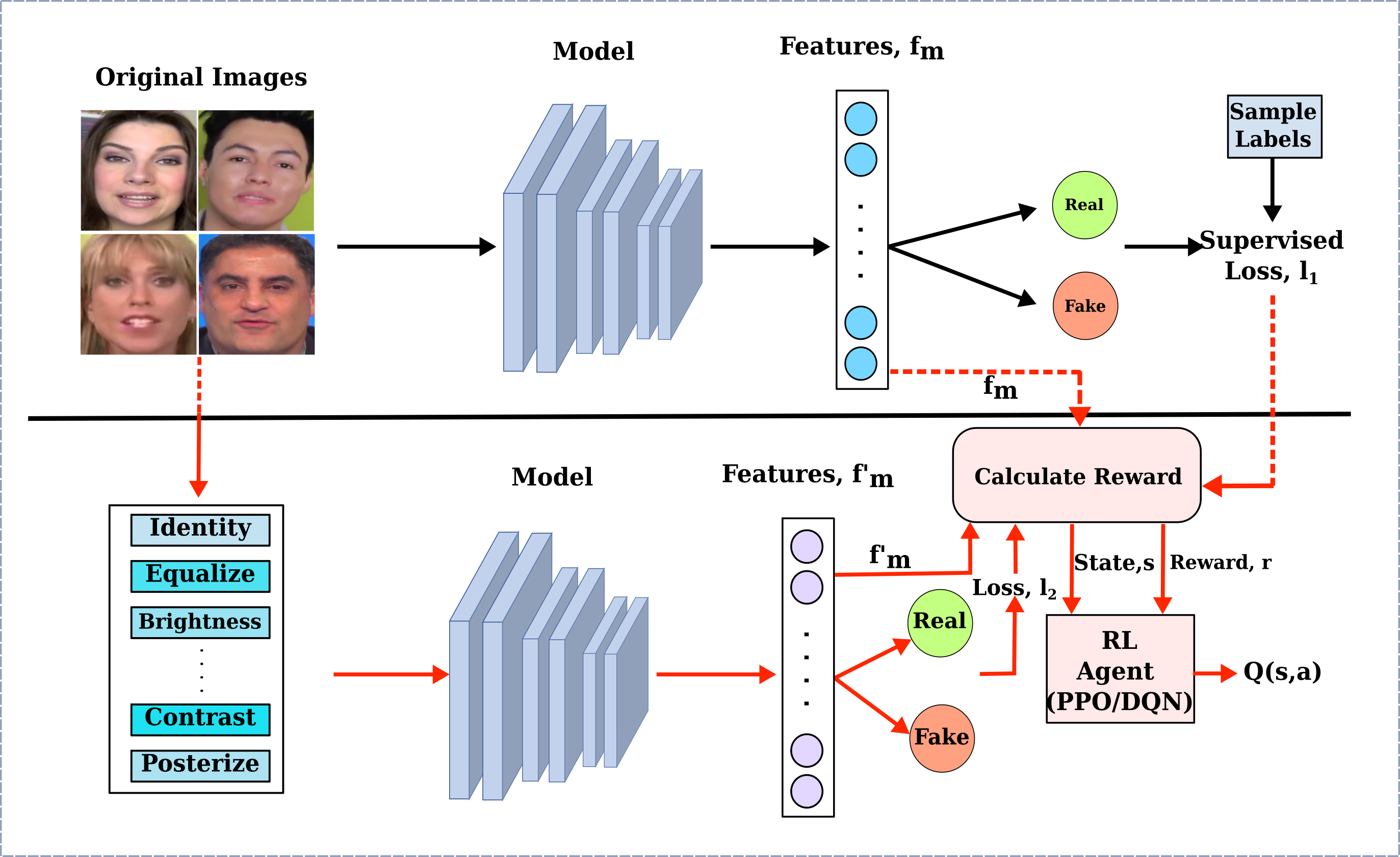}}
\caption{Illustration of the combination of supervised and deep reinforcement learning for training the RL agent for top-$k$ test-time augmentations selection based on the learned policy. 
} 
\label{fig}
\end{figure*}

To mitigate the risk posed by deep fakes, the popular deep fake detection methods include training
convolutional neural networks (CNNs) (such as ResNet-50~\cite{He2016DeepRL}, XceptionNet~\cite{8099678}, and InceptionNet~\cite{7780677}) for detecting facial manipulations from real imagery. These CNNs are trained to detect visual artifacts or blending boundary of a forged image. The deep fake datasets such as Celeb-DF~\cite{9156368}, FaceForensics++~\cite{9010912}, DeeperForensics-$1.0$~\cite{jiang2020deeperforensics1} and DFDC~\cite{https://doi.org/10.48550/arxiv.2006.07397} have been assembled for research and development in detecting deep fakes. These datasets can be categorized into different generations depending on the quality and the amount of the real and fake data.

Most of the aforementioned CNN-based deep fake detection methods obtain very high performance in the intra-dataset evaluation (i.e., when the same dataset is used for training and testing). However, they obtain poor generalization across datasets. For instance, XceptionNet for deep fake detection obtained Area Under Curve (AUC) of $0.997$ when trained and tested on FaceForensics++ dataset. However, the AUC score reduced to $0.482$ when tested on Celeb-DF in cross-dataset scenario~\cite{8099678}. The degradation in the performance across datasets is due to \textbf{domain shift} i.e., the data distribution change between the training and testing set. This is due to change in the image quality of real videos and deep fakes following the continuous advances in sensor technology and the deep fake generation techniques. 

The aim of this paper is to \emph{improve the cross-dataset generalization} of the current CNN-based deep fake detectors. To this front, a hybrid combination of the supervised and deep reinforcement learning (RL) module is proposed. 
Reinforcement learning (RL)~\cite{vanOtterlo2012,Arulkumaran} is a branch of machine learning that studies how an intelligent agent should operate in a given environment to maximize the concept of cumulative reward. In a standard reinforcement learning model, the agent receives an input, which is the current state of the environment. The agent then chooses an action that changes the current state and the value of the state transition is communicated to the agent through a scalar reinforcement signal called reward. Through systematic trial and error, the agent learns a policy to choose actions that tend to maximize the cumulative reward.

Our proposed approach can meticulously choose and apply top-$k$ augmentations to the test samples via a learned policy by an RL agent in an image-specific manner. The scores of the augmented test samples are averaged together for the final classification. Our experimental results demonstrate the \emph{merit} of test-time image-specific augmentations in \emph{reducing the impact of domain shift} in cross-dataset performance evaluation of deep fake detectors. Experimental investigation on FaceForensics++, Celeb-DF, and DeeperForensics-$1.0$ show that our approach obtains state-of-the-art performance in cross-dataset generalization of the deep fake detectors over other published work.

In summary, the major \textbf{contributions} of this paper are threefold as given below:

\begin{itemize}
    \item We propose a hybrid combination of supervised and reinforcement learning techniques that applies top-$k$ augmentations to the test samples in an image-specific manner using a policy learned by an RL agent, for deep fake classification. 
    
    \item We demonstrate the merit of our approach in reducing the impact of domain shift over random test-time data augmentations in performance evaluation of deep fake detectors across datasets. 
    
    \item Extensive experiments demonstrate the efficacy of our approach in improving the cross-dataset generalization of deep fake detection over published work, thus obtaining state-of-the-art performance in deep fake detection across datasets over existing classification baselines.
    
\end{itemize}

This paper is organized as follows: Section~\ref{section2} discusses the prior work on deep fake detection. Section~\ref{section3} describes our proposed method for training the RL agent and choosing the top-$k$ augmentations during the test-time for deep fake classification. Datasets and experimental protocol are discussed in section~\ref{section4}. Experimental results are discussed in section~\ref{section5}. The ablation study for choosing the top-$k$ augmentations is detailed in section~\ref{section6}. Conclusion and future work are discussed in section~\ref{section7}.

\section{Prior Work on Deepfake Detection}
\label{section2}
In this section, we will discuss the existing countermeasure proposed for deep fake detection. 
Most of the existing methods are CNN-based classification baselines trained for deep fake detection~\cite{He2016DeepRL,8099678,Nguyen2019MultitaskLF,9717407}.

In ~\cite{Li_2019_CVPR_Workshops}, Li and Lyu used VGG16, ResNet50, ResNet101, and ResNet152 based CNNs for the detection of the presence of artifacts from the facial regions and the surrounding areas for deep fake detection. 
Afchar et al.~\cite{8630761} proposed two different CNN architectures composed of only a few layers in order to focus on the mesoscopic properties of the images: (a) a CNN comprised of 4 convolutional layers followed by a fully-connected layer (Meso-4), and (b) a modification of Meso-4 using a variant of the Inception module named MesoInception-4. 
Zhou et al.~\cite{8014963} proposed a two-stream network for face manipulation detection. In particular, the authors considered a fusion of a face classification stream based on the CNN GoogLeNet and a path triplet stream that is trained using steganalysis features of images patches. In~\cite{9010912}, an exhaustive analysis of different CNN-based deep fake detection methods by Rosslet et al. suggested efficacy of XceptionNet when evaluated on FaceForensics++. 

Nguyen et al.~\cite{Nguyen2019MultitaskLF} proposed a multi-task CNN to simultaneously detect the fake videos and locate the manipulated regions using an autoencoder with a Y-shaped decoder for information sharing between classification, segmentation, and reconstruction tasks.
In~\cite{9717407}, biometric tailored loss functions (such as Center, ArcFace, and A-Softmax) are used for two-class CNN training for deep fake detection. 
In~\cite{9157215}, a face X-ray model has been proposed to detect forgery by detecting the blending boundary of a forged image using a two-class CNN model trained end-to-end.


Apart from the aforementioned CNN-based deep fake detection methods, facial  and behavioral biometrics (i.e., facial expression, head, and body movement) have been used for deep fake detection~\cite{Dong2020IdentityDrivenDD, 9360904, agarwal_protecting_2019, 9717407}. 
Study in~\cite{10.1007/978-3-030-58571-6_39} used a two-branch representation extractor that combines information from the color and the frequency domain using a multi-scale Laplacian of Gaussian (LOG) operator.
Very recently in~\cite{9577592}, a multi-attentional deep fake detection based on multiple spatial attention heads along with feature aggregation guided by the attention maps is proposed. 

Readers are referred to published survey in~\cite{Tolosana2020deepfakesAB},~\cite{Nguyen2019DeepLF} for detailed information on deep fake detection methods.

\section{Proposed Method}
\label{section3}
In this section, we provide a detailed explanation of the proposed method which works as follows: during the \emph{training stage}, the deep CNN is trained on live and fake images for deep fake classification. Using the hybrid combination of 
 supervised learning (i.e., the CNN output) and reinforcement learning, RL agent is trained for optimum augmentations (action) selection in an image-specific manner. 
  
  Figure~\ref{fig} illustrates the steps involved in training the RL agent for optimum action selection. For each training sample, the CNN outputs the deep feature map ($f_m$) and the associated loss ($l_1$) which is the cross-entropy loss. This is followed by the reinforcement learning module.  A typical RL technique is as follows: A RL agent receives reward $r$ from the environment along with the current state $s$. Depending on the inputs ($r$,~$s$), the RL agent learns a policy to select the right action $a$ for the environment that maximizes the cumulative reward and finally generates the Q-table containing the maximum expected reward for each state, action ($s$,~$a$) pairs~\cite{10.5555/1622737.1622748}~\cite{vanOtterlo2012}.

In the context of our application, the RL module is composed of seven main components namely:
\begin{itemize}
    \item \textbf{Environment}: The CNN model
    \item \textbf{State (s)}: Feature map ($f_m$) obtained from CNN model
    \item \textbf{Action (a)}: Data augmentations
    \item \textbf{RL agent}: RL agent takes state, reward  as input and generates state-action pairs
    \item \textbf{Reward (r)}: Reward plays a key role in adjusting agents policy $\Pi_\Theta(a,s)$
    \item \textbf{Policy ($\Pi_\Theta(a,s)$)}: Useful for calculating Q-table
    \item \textbf{Q-table} (Q(s,a)): It contains values for state-action pairs
  \end{itemize}

An  action ($a$) is chosen from a bank of ten augmentations. The training image is then subjected to the action permutation. The permuted image's new feature map $f'_{m}$ is then extracted from the CNN along with the associated loss, $l_{2}$. On comparing loss values $l_{1}$,$l_{2}$ of the selected image before and after applying the action permutation, the current state (feature map) and the reward ($r$) is obtained using equation~\ref{my_fifth_eqn}. As given in equation~\ref{my_fifth_eqn}, if $l_{2}<l_{1}$ the current state will be $f^{'}_{m}$ and the reward will be $l_{1}$-$l_{2}$. Otherwise, if $l_2\geq l_{1}$ the current state will be $f_{m}$ and the reward will be calculated as $l_{1}$-$l_{2}$.

\begin{equation}\label{my_fifth_eqn}
reward(r)=\begin{Bmatrix}\\
l_{1}-l_{2},& if & l_2< l_1\rightarrow f^{'}_{m}\\\\
l_{2}-l_{1} ,& if & l_2\geq l_{1}\rightarrow f_{m}\\
 &  &
\end{Bmatrix}
\end{equation}
This process is repeated for each training sample and the policy $\pi_\theta(a,s)$ is learned by the RL agent based on the state ($s$) of the environment and the calculated reward ($r$). 
The final output of the RL agent is a Q-table, $Q(s, a)$, which contain cumulative reward for state-action pairs based on the learned policy $\Pi_\Theta(a,s)$.

During the \emph{testing stage}, the feature map of the test sample obtained by the CNN is given as an input (current state $s$) to the RL agent. The RL agent outputs the scores from Q-table for each action  (augmentation) based on the learned policy. The top-$k$ augmentations are applied to the test sample based on the scores obtained from the Q-table. The classification scores obtained from the trained CNN for each $k^{th}$ augmentation of the test sample are averaged together for the final real/fake classification. 

In our experiments, we used $10$ augmentations namely, ’Identity’, ’AutoContrast’, ’Equalize’,’ Rotate’, ’Solarize’, ’Color’, ’Posterize’,’ Contrast’, ’Brightness’, ’Sharpness’,’ ShearX’, ’ShearY’, ’TranslateX’, ’TranslateY’ as the actions (a) to be performed by the RL agent. We evaluated Proximal Policy Optimization~(PPO)~\cite{https://doi.org/10.48550/arxiv.1707.06347} and Deep Q Network (DQN)~\cite{Mnih2013PlayingAW} based RL agents for learning the policy and choosing the right actions~$(a)$. These RL agents are explained below.



\subsection{Proximal Policy Optimization (PPO)}

\begin{figure}[htbp]
\centerline{\includegraphics[width=0.48\textwidth]{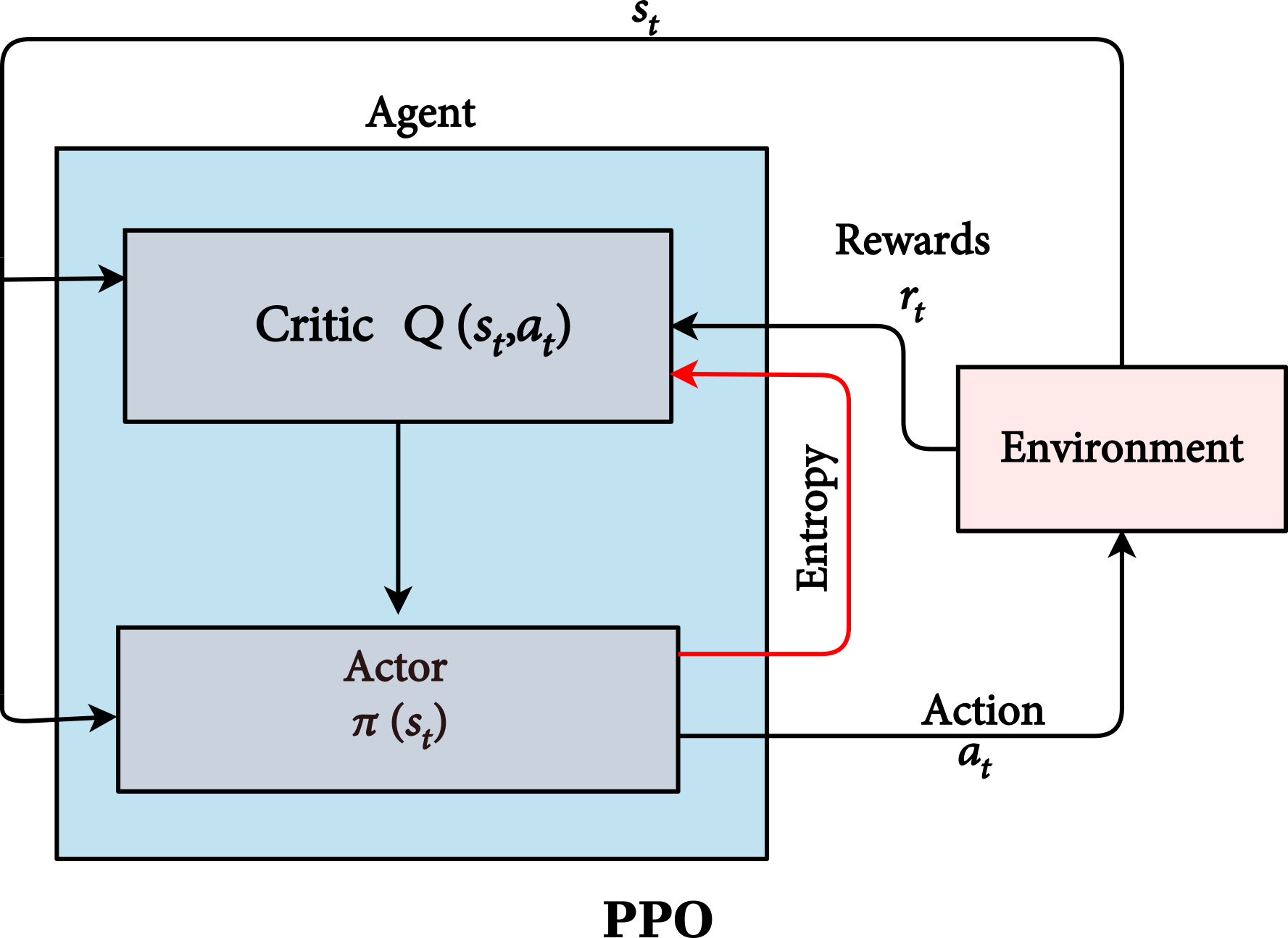}}
\caption{Illustration of the Proximal Policy Optimization (PPO) based RL agent. PPO is based on actor and critic models which helps to generate a policy gradient and a Q-table for the environment. The actor and critic components are helpful to learn the policy $\pi_\theta(s_{t})$ and generate the $Q(s_{t},a_{t})$. An action $a_{t}$ is selected for the environment by the PPO  by time $t$.}
\label{fig1}
\end{figure}

In PPO algorithm~\cite{https://doi.org/10.48550/arxiv.1707.06347}, there is an actor and a critic model as illustrated in Figure~\ref{fig1}. The role of the actor corresponds to the policy $\pi$ and is used to choose the action for the agent and update the policy network. The critic corresponds to the value function $Q(s,a)$ for an  action value or $V(s)$ for a state value. PPO policy would be learned by calculating an estimator of the policy gradient and using it with a stochastic gradient descent algorithm for approximating the function. The most widely used gradient estimator is defined as:
\begin{equation}\label{my_sixth_eqn}
\hat{g} = \mathbb{\hat{E}}_{t}\left [ \bigtriangledown _{\Theta }\log \pi _{\Theta} \left ( a_{t}|s_{t} \right ) \hat{A}_{t}\right ]
\end{equation}\\
where $\pi_\theta$ is a stochastic policy and $\hat{A}_{t}$ is the estimator of the advantage function at timestep $t$. Here $\mathbb{\hat{E}}_{t}$  is the empirical average over a finite batch of samples. The objective function is determined with the help of this gradient estimator and the aim of objective function is to maximize the size of policy update. Finally, the updated policy is used to predict the action ($a$) which is given as an input to the model or environment. Among all the objective functions, the best surrogate objective function is $L^{CLIP}_t(\theta)$ which obtained highest average normalized score for the PPO algorithm over other surrogate objectives. Therefore, we used $L^{CLIP}_t(\theta)$ for this study given in~\cite{https://doi.org/10.48550/arxiv.1707.06347}.

\subsection{Deep Q Network (DQN)}
\begin{figure}[htbp]
\centerline{\includegraphics[width=0.48\textwidth]{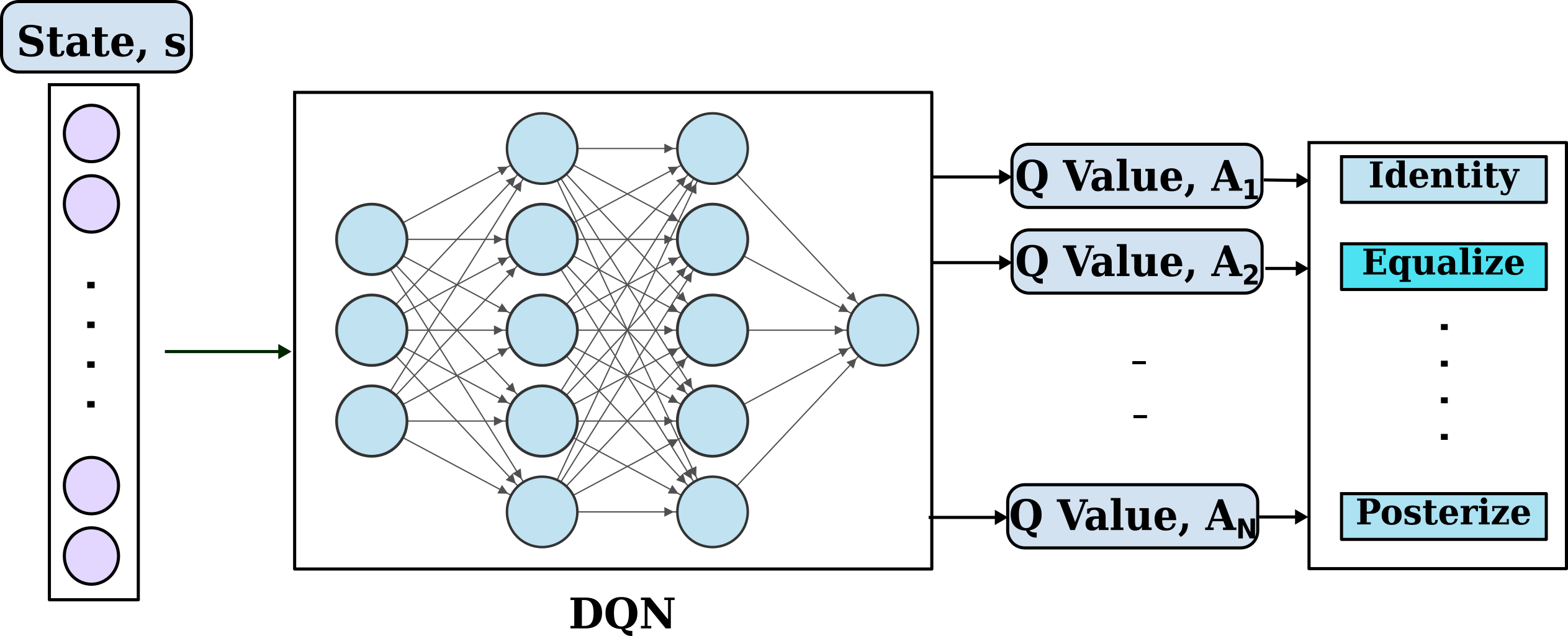}}
\caption{Illustration of the Deep Q Network (DQN) based RL approach which uses a neural network to approximate the Q-value function. The state ($s$) is given as an input, and the output is the Q-value of all potential actions ($a$). In this study, state $s$ is the feature map and the set of actions ($a$) are the set of augmentations ($A_i$).}
\label{fig2}
\end{figure}

In Deep Q Network learning (DQN)~\cite{Mnih2013PlayingAW}, a neural network is learned to approximate the Q-value function as seen in Figure~\ref{fig2}. The state (s) is given as an input and the output is the Q-value of all potential actions (which are the set of augmentations in this study). 
The optimal action-value function obeys an important identity known as the Bellman equation. A
Deep Q-network~\cite{Mnih2013PlayingAW} can be trained by minimizing a sequence of loss functions $L_i(\theta_i)$ that changes at each
iteration $i$ and the objective function is defined as:

\begin{equation}\label{my_eighth_eqn}
L_{i}\left ( \theta_{i} \right )=\mathbb{E}_{s,a\sim \rho (.)} \left [ \left ( y_{i}-Q\left ( s,a;\theta_{i}\right ) \right )^{2} \right ]
\end{equation}

where $y_{i}=\mathbb{E}_{s^{‘}\sim \varepsilon }\left [ r+\gamma max_{a^{‘}}Q(s^{‘},a^{’};\theta_{i-1} )|s,a \right ]$  is the target for iteration $i$ and $\rho(s )$ is a
probability distribution over state $s$ and actions $a$ that is referred to as the behaviour distribution. On differentiating the objective function, the policy gradient $\pi_\theta$ is obtained to learn a policy that uses the optimal strategy to select the optimal action for maximizing the reward.
The final Q-table will be a function of $Q^*(s,a)$ which is defined as:
\begin{equation}\label{my_ninth_eqn}
Q^{*}\left ( s,a \right )= \mathbb{E}_{s^{‘}\sim \varepsilon }\left [ r+\gamma maxQ^{*}(s^{‘},a^{’})|s,a \right ]
\end{equation}

The $\gamma$ is a hyperparameter which is a fixed value and $r$ is the reward that needs to be maximized. In this study, $\gamma$ is set to $0.5$ based on empirical evidence.

\section{Dataset and Experimental Protocol}
\label{section4}

In all the experiments, we used state-of-the-art FaceForensics++~\cite{9010912}, Deeper Forensics-$1.0$~\cite{jiang2020deeperforensics1} and Celeb-DF~\cite{9156368} datasets. These datasets are discussed as follows:

\begin{itemize}
  \item \textbf{FaceForensics++:} 
  FaceForensics++~\cite{9010912} is an automated benchmark for facial manipulation detection. It consists of several manipulated videos created using two different generation techniques: Identity Swapping (FaceSwap, FaceSwap-Kowalski, FaceShifter, deep fakes) and Expression swapping (Face2Face and NeuralTextures). 
 We used the FaceForensics++ dataset's $c23$  version for both training and testing, which has a curated list of $70$ videos for each of these deep fake creation methods.
 \item \textbf{Celeb-DF:} 
 The Celeb-DF~\cite{9156368} deep fake forensic dataset include $590$ genuine videos from $59$ celebrities as well as $5639$ deep fake videos. Celeb-DF, in contrast to other datasets, has essentially no splicing borders, color mismatch, and inconsistencies in face orientation, among other evident deep fake visual artifacts. The deep fake videos in Celeb-DF are created using an encoder-decoder style model which results in better visual quality. 
  \item \textbf{DeeperForensics-1.0:} 
  The DeeperForensics-$1.0$~\cite{jiang2020deeperforensics1} is one of the largest deep fake dataset used for face forgery detection. DF-1.0 consists of $60,000$ videos that have around $17.6$ million frames with substantial real-world perturbations. The dataset contains videos of $100$ consented actors with $35$ different perturbations. The real to fake videos ratio is $5$:$1$ and the fake videos are generated by an end-to-end face-swapping framework. 

\begin{table*}
\caption {Evaluation of the existing CNN-based deep fake detection baselines trained on FaceForensics++ and tested on FaceForensics++, DeeperForensics-$1.0$, and Celeb-DF. In order to establish the merit of our approach of systematic selection of augmentations for test samples by an RL agent, we also evaluated the current deep fake detectors with random test-time augmentations. The top performance results are highlighted in bold.}
\label{Table1}
\begin{center}
\scalebox{0.8}{
\begin{tabular}{c|c|l|lllllllll}
\hline
                               & \multicolumn{1}{l|}{}                                                 &                                    & \multicolumn{9}{c}{Evaluation Datasets}                                                                                                                                                                                                                                                                                        \\ \cline{4-12} 
                               & \multicolumn{1}{l|}{}                                                 &                                    & \multicolumn{3}{c|}{FaceForensics++}                                                                            & \multicolumn{3}{c|}{DeeperForensics-1.0}                                                                        & \multicolumn{3}{c}{Celeb-DF}                                                               \\ \cline{4-12} 
\multirow{-3}{*}{Models}       & \multicolumn{1}{l|}{\multirow{-3}{*}{Test time Augmentation}}              & \multirow{-3}{*}{Training Dataset} & \multicolumn{1}{l|}{AUC}            & \multicolumn{1}{l|}{pAUC}           & \multicolumn{1}{l|}{EER}            & \multicolumn{1}{l|}{AUC}            & \multicolumn{1}{l|}{pAUC}           & \multicolumn{1}{l|}{EER}            & \multicolumn{1}{l|}{AUC}            & \multicolumn{1}{l|}{pAUC}           & EER            \\ \hline
ResNet-50                      & -& FF++                               & \multicolumn{1}{l|}{0.945}          & \multicolumn{1}{l|}{0.908}          & \multicolumn{1}{l|}{0.125}          & \multicolumn{1}{l|}{0.885}          & \multicolumn{1}{l|}{0.843}          & \multicolumn{1}{l|}{0.208}          & \multicolumn{1}{l|}{0.625}          & \multicolumn{1}{l|}{0.584}          & 0.405          \\ \hline
XceptionNet                   & -                                                                    & FF++                               & \multicolumn{1}{l|}{0.985}          & \multicolumn{1}{l|}{0.969}          & \multicolumn{1}{l|}{0.037}          & \multicolumn{1}{l|}{\textbf{0.940}} & \multicolumn{1}{l|}{\textbf{0.825}} & \multicolumn{1}{l|}{\textbf{0.127}} & \multicolumn{1}{l|}{0.651}          & \multicolumn{1}{l|}{0.629}          & 0.383          \\ \hline
EfficientNet V2-L              & -                                                                     & FF++                               & \multicolumn{1}{l|}{\textbf{0.991}} & \multicolumn{1}{l|}{\textbf{0.979}} & \multicolumn{1}{l|}{\textbf{0.024}} & \multicolumn{1}{l|}{0.938}          & \multicolumn{1}{l|}{0.882}          & \multicolumn{1}{l|}{0.142}          & \multicolumn{1}{l|}{\textbf{0.658}} & \multicolumn{1}{l|}{\textbf{0.635}} & \textbf{0.379} \\ \hline
InceptionNet                  & -                                                                     & FF++                               & \multicolumn{1}{l|}{0.922}          & \multicolumn{1}{l|}{0.852}          & \multicolumn{1}{l|}{0.187}          & \multicolumn{1}{l|}{0.859}          & \multicolumn{1}{l|}{0.826}          & \multicolumn{1}{l|}{0.225}          & \multicolumn{1}{l|}{0.598}          & \multicolumn{1}{l|}{0.552}          & 0.459          \\ \hline
   {ResNet-50} & \checkmark                                                              & FF++                               & \multicolumn{1}{l|}{0.919}          & \multicolumn{1}{l|}{0.882}          & \multicolumn{1}{l|}{0.193}          & \multicolumn{1}{l|}{0.849}          & \multicolumn{1}{l|}{0.826}          & \multicolumn{1}{l|}{0.248}          & \multicolumn{1}{l|}{0.592}          & \multicolumn{1}{l|}{0.564}          & 0.479          \\ \hline
XceptionNet                   & \checkmark                                                           & FF++                               & \multicolumn{1}{l|}{0.959}          & \multicolumn{1}{l|}{0.927}          & \multicolumn{1}{l|}{0.124}          & \multicolumn{1}{l|}{\textbf{0.912}} & \multicolumn{1}{l|}{\textbf{0.875}} & \multicolumn{1}{l|}{\textbf{0.197}} & \multicolumn{1}{l|}{0.615}          & \multicolumn{1}{l|}{0.589}          & 0.392          \\ \hline
EfficientNet V2-L              & \checkmark                                                            & FF++                               & \multicolumn{1}{l|}{\textbf{0.967}} & \multicolumn{1}{l|}{\textbf{0.946}} & \multicolumn{1}{l|}{\textbf{0.118}} & \multicolumn{1}{l|}{0.907}          & \multicolumn{1}{l|}{0.864}          & \multicolumn{1}{l|}{0.187}          & \multicolumn{1}{l|}{\textbf{0.622}} & \multicolumn{1}{l|}{\textbf{0.598}} & \textbf{0.385} \\ \hline
InceptionNet                  & \checkmark                                                            & FF++                               & \multicolumn{1}{l|}{0.896}          & \multicolumn{1}{l|}{0.874}          & \multicolumn{1}{l|}{0.195}          & \multicolumn{1}{l|}{0.828}          & \multicolumn{1}{l|}{0.814}          & \multicolumn{1}{l|}{0.275}          & \multicolumn{1}{l|}{0.569}          & \multicolumn{1}{l|}{0.545}          & 0.495          \\ \hline
\end{tabular}}

\end{center}
\end{table*}
\end{itemize}

\begin{table*}[]
\caption{Evaluation of our proposed model in intra and cross-dataset scenarios with PPO and DQN as RL Agents. The top-$3$ augmentations are selection for each test sample. The classification scores from top-$3$ augmentations of a test sample are averaged for final deep fake detection. The top performances are highlighted in bold.}
\label{table2}
\begin{center}    

\scalebox{0.90}{
\begin{tabular}{c|cc|l|lllllllll}
\hline
\multirow{3}{*}{Models} & \multicolumn{2}{l|}{RL Methods}                                                       & \multirow{3}{*}{Training Dataset} & \multicolumn{9}{c}{Evaluation Datasets}                                                                                                                                                                                                                                                                                        \\ \cline{2-3} \cline{5-13} 
                        & \multicolumn{1}{l|}{\multirow{2}{*}{PPO}} & \multicolumn{1}{l|}{\multirow{2}{*}{DQN}} &                                   & \multicolumn{3}{c|}{FaceForensics++}                                                                            & \multicolumn{3}{c|}{DeeperForensics-1.0}                                                                        & \multicolumn{3}{c}{Celeb-DF}                                                               \\ \cline{5-13} 
                        & \multicolumn{1}{l|}{}                     & \multicolumn{1}{l|}{}                     &                                   & \multicolumn{1}{l|}{AUC}            & \multicolumn{1}{l|}{pAUC}           & \multicolumn{1}{l|}{EER}            & \multicolumn{1}{l|}{AUC}            & \multicolumn{1}{l|}{pAUC}           & \multicolumn{1}{l|}{EER}            & \multicolumn{1}{l|}{AUC}            & \multicolumn{1}{l|}{pAUC}           & EER            \\ \hline
ResNet-50               & \multicolumn{1}{c|}\checkmark       & -                          & FF++                              & \multicolumn{1}{l|}{0.948}          & \multicolumn{1}{l|}{0.914}          & \multicolumn{1}{l|}{0.165}          & \multicolumn{1}{l|}{0.897}          & \multicolumn{1}{l|}{0.854}          & \multicolumn{1}{l|}{0.203}          & \multicolumn{1}{l|}{0.629}          & \multicolumn{1}{l|}{0.592}          & 0.398          \\ \hline
Xception Net            & \multicolumn{1}{c|}\checkmark       & -                                         & FF++                              & \multicolumn{1}{l|}{0.989}          & \multicolumn{1}{l|}{0.972}          & \multicolumn{1}{l|}{0.035}          & \multicolumn{1}{l|}{0.944}          & \multicolumn{1}{l|}{0.912}          & \multicolumn{1}{l|}{0.125}          & \multicolumn{1}{l|}{0.657}          & \multicolumn{1}{l|}{0.625}          & 0.376          \\ \hline
Efficient Net V2-L       & \multicolumn{1}{c|}\checkmark     & -                                         & FF++                              & \multicolumn{1}{l|}{\textbf{0.994}} & \multicolumn{1}{l|}{\textbf{0.983}} & \multicolumn{1}{l|}{\textbf{0.022}} & \multicolumn{1}{l|}{\textbf{0.952}} & \multicolumn{1}{l|}{\textbf{0.922}} & \multicolumn{1}{l|}{\textbf{0.119}} & \multicolumn{1}{l|}{\textbf{0.669}} & \multicolumn{1}{l|}{\textbf{0.647}} & \textbf{0.362} \\ \hline
Inception Net           & \multicolumn{1}{c|}\checkmark             & -                                         & FF++                              & \multicolumn{1}{l|}{0.935}          & \multicolumn{1}{l|}{0.886}          & \multicolumn{1}{l|}{0.148}          & \multicolumn{1}{l|}{0.869}          & \multicolumn{1}{l|}{0.847}          & \multicolumn{1}{l|}{0.218}          & \multicolumn{1}{l|}{0.608}          & \multicolumn{1}{l|}{0.576}          & 0.424          \\ \hline
ResNet-50               & \multicolumn{1}{c|}{-}                    & \checkmark                                 & FF++                              & \multicolumn{1}{l|}{0.937}          & \multicolumn{1}{l|}{0.912}          & \multicolumn{1}{l|}{0.139}          & \multicolumn{1}{l|}{0.879}          & \multicolumn{1}{l|}{0.834}          & \multicolumn{1}{l|}{0.253}          & \multicolumn{1}{l|}{0.619}          & \multicolumn{1}{l|}{0.572}          & 0.419          \\ \hline
Xception Net            & \multicolumn{1}{c|}{-}                    & \checkmark                                 & FF++                              & \multicolumn{1}{l|}{0.979}          & \multicolumn{1}{l|}{0.952}          & \multicolumn{1}{l|}{0.039}          & \multicolumn{1}{l|}{0.937}          & \multicolumn{1}{l|}{0.898}          & \multicolumn{1}{l|}{0.129}          & \multicolumn{1}{l|}{\textbf{0.656}} & \multicolumn{1}{l|}{\textbf{0.642}} & \textbf{0.356} \\ \hline
Efficient Net V2-L       & \multicolumn{1}{c|}{-}                    & \checkmark                                 & FF++                              & \multicolumn{1}{l|}{\textbf{0.982}} & \multicolumn{1}{l|}{\textbf{0.965}} & \multicolumn{1}{l|}{\textbf{0.039}} & \multicolumn{1}{l|}{\textbf{0.948}} & \multicolumn{1}{l|}{\textbf{0.917}} & \multicolumn{1}{l|}{\textbf{0.122}} & \multicolumn{1}{l|}{0.647}          & \multicolumn{1}{l|}{0.628}          & 0.385          \\ \hline
Inception Net           & \multicolumn{1}{c|}{-}                    & \checkmark                            & FF++                              & \multicolumn{1}{l|}{0.927}          & \multicolumn{1}{l|}{0.899}          & \multicolumn{1}{l|}{0.184}          & \multicolumn{1}{l|}{0.854}          & \multicolumn{1}{l|}{0.822}          & \multicolumn{1}{l|}{0.229}          & \multicolumn{1}{l|}{0.592}          & \multicolumn{1}{l|}{0.549}          & 0.495          \\ \hline
\end{tabular}}

\end{center}
\end{table*}

\noindent \textbf{Experimental Protocol:}
We evaluated four different CNN architectures namely, ResNet-50~\cite{He2016DeepRL}, InceptionNet-v3~\cite{7780677}, EfficientNet v2-L~\cite{pmlr-v139-tan21a} and XceptionNet~\cite{8099678} for deep fake detection. These models were trained on FaceForensics++ $c23$ version which is a high quality (HQ) version of FF++. The
face images were detected and aligned using MTCNN~\cite{7553523}.
MTCNN utilizes a cascaded CNNs based framework for joint face detection and alignment. 
The images are then resized to $256\times256$ for both training and evaluation.  

For all the CNN models, we used a batch-normalization layer followed by the last fully connected layer of size $1024$ and the final output layer for deep fake classification. The CNN models are trained using an Adam optimizer with an initial learning rate of $0.001$ and a weight decay of 1e6. The models are trained on $4$ RTX $5000$Ti GPUs with a batch size of $64$. The same training set is used for training the RL agents for learning the optimum policy using our proposed model. 

We used the sampling approach described in~\cite{9010912} to choose $270$ frames per video for training and $150$ frames per video for validation and testing the models. 
The trained models are tested on FaceForensics++(HQ), Deeper Forensics-$1.0$, and Celeb-DF datasets.  
The standard performance metrics used for deep fake detection namely, Area under the Curve~(AUC), Partial Area under the Curve~(pAUC) at $10\%$ False Positive Rate~(FPR), and the Equal Error Rate~(EER) are computed at frame level for the evaluation.

\section{Results and Discussion}
\label{section5}
In this section, we discuss the CNN-based deep fake classification baselines in the intra- and cross-dataset evaluation with and without random test-time augmentations in subsection~\ref{subsection51}. Evaluation results of our proposed hybrid model using PPO and DQN based RL agents in the intra- and cross-dataset scenario are discussed in subsection~\ref{subsection52}. In subsection~\ref{subsection53}, we compare the performance of the existing published results on deep fake detection across datasets with our best results. 


\subsection{Cross-dataset Generalization of CNN-based Deepfake Detectors}
\label{subsection51}
Table~\ref{Table1} shows the performance of the CNN-based deep fake detectors trained on FaceForensics++ and tested on FaceForensics++, DeeperForensics-$1.0$ and Celeb-DF. These performances are reported with and without test time augmentations. For test-time augmentations, three random augmentations are applied to each test sample and the classification scores are averaged for deep fake detection. 

The top performance results are highlighted in bold across various evaluation datasets. Mostly, Efficient v2-L obtained the best results with an AUC of $0.991$ and EER of $0.024$ when trained and evaluated on FaceForensics++ (intra-dataset). 
On average, AUC, pAUC, and EER of $0.960$, $0.927$, and $0.093$, respectively, were obtained across the models in intra-dataset evaluation without test-time augmentations.

Across datasets, the performance of all the \emph{models} dropped significantly.
On average, AUC, pAUC, and EER of $0.905$, $0.844$, and $0.175$  were obtained across the models when trained on FaceForensics++ and evaluated on DeeperForensics-$1.0$. 
On the Celeb-DF dataset, on average, AUC, pAUC, and EER of $0.633$, $0.60$, and $0.406$, respectively, were obtained across the models when trained on FaceForensics++ and evaluated on Celeb-DF. 

Mostly, the EfficientNet V2-L backbone obtained the best performance in the intra- and cross-dataset. The \emph{performance drop of the models across datasets is significant for Celeb-DF over DeeperForensics-$1.0$}. The reason being deep fake videos in Celeb-DF are created using an encoder-decoder style model which results in better visual quality. Whereas, the videos created in DeeperForensics-$1.0$ uses deep fake generation techniques similar to FaceForenics++.

For most of the models, random test-time augmentations reduced their performance in intra- and cross-dataset evaluation. \emph{This suggests the need for systematic selection of test-time augmentations in an image-specific manner for an enhanced performance}.




\subsection{Evaluation of Our Proposed Model}
\label{subsection52}

Table~\ref{table2} shows the deep fake detection performance of our proposed model for both PPO and DQN based RL agents. The top performance results are highlighted in bold across various datasets. All the results are reported for top-$3$ augmentations applied to the test samples selected by an RL agent before deep fake classification. The top-$3$ augmentations are selected based on empirical evidence.

As can be seen from the table, EfficientNet V2-L along with PPO-based RL agent is consistently the best performing model when evaluated across different datasets.
This model with PPO obtained an AUC of $0.994$ and an EER of $0.022$ when trained and evaluated on FaceForensics++.

On average, AUC, pAUC, and EER of $0.966$, $0.938$, and $0.092$ were obtained across the models with PPO-based RL agent in the intra-dataset evaluation. \emph{This suggests that intra-dataset performance of the CNN-based deep fake detectors remain intact when top-$k$ augmentations are applied to test samples by an RL agent using our proposed model}.

\begin{table}
\caption{Comparison with existing studies on cross dataset evaluation of deep fake detection on Celeb-DF. The results from existing methods are directly taken from~\cite{10.1007/978-3-030-58571-6_39}. AUC scores are used for comparison.}
\label{table3}
\begin{center}
\scalebox{0.89}{

\begin{tabular}{l|l|c}
\hline
Method                            & FF++  & \multicolumn{1}{l}{Celeb-DF} \\ \hline
Mes04~\cite{8630761}                             & 0.847  & 0.548                        \\
FWA~\cite{Li_2019_CVPR_Workshops}                               & 0.801 & 0.569                        \\
Xception-raw~\cite{9156368}                       & 0.997 & 0.482                        \\
Xception-c23~\cite{9156368}                       & 0.997 & 0.653                        \\
Xception-c40~\cite{9156368}                       & 0.955 & 0.655                        \\
MesoInception4~\cite{8630761}                     & 0.830 & 0.536                        \\
Two-stream~\cite{8014963}                       & 0.701 & 0.538                        \\
Multi-task~\cite{Nguyen2019MultitaskLF}                         & 0.763 & 0.543                        \\
Capsule~\cite{9156368}                            & 0.966 & 0.575                        \\
DSP-FWA~\cite{Li_2019_CVPR_Workshops}                            & 0.930 & 0.646                        \\
Two Branch~\cite{10.1007/978-3-030-58571-6_39}                         & 0.931 & 0.734                        \\

$F^{3}$-Net~\cite{10.1007/978-3-030-58610-2_6}                        & 0.981 & 0.651                        \\ EfficientNet-B4~\cite{pmlr-v97-tan19a}                    & 0.997 & 0.642                        \\\hline
\textbf{Ours (EfficientNet-v2-L with PPO)} & \textbf{0.994} & \textbf{0.669}                        \\ \hline
\end{tabular}
}
\end{center}
\end{table}

Across datasets, EfficientNet v2-L with PPO-based RL agent has obtained the highest AUC and the least EER of $0.952$ and $0.119$, respectively, when evaluated on DeeperForensics-$1.0$. 
When evaluated on Celeb-DF, EfficientNet v2-L with PPO-based RL agent has obtained the highest AUC and the least EER of $0.669$ and $0.362$, respectively. 
On average, AUC, pAUC, and EER of $0.640$, $0.610$, and $0.390$ were obtained across the models with the PPO-based RL agent when evaluated on Celeb-DF. 

The PPO-based RL agent outperformed DQN for all the models. The reason is as an on-policy algorithm, PPO tackles the problem of sample efficiency by employing surrogate objectives to keep the new policy from drifting too much from the old policy~\cite{https://doi.org/10.48550/arxiv.1707.06347}. This aids in learning the optimal policy. \\

\noindent \textbf{Comparison with the baseline CNNs in Table~\ref{Table1}}: 
In comparison to the baseline in Table~\ref{Table1}, all the same CNN architectures along with an RL agent for optimum test-time augmentations selection, obtained performance improvement across datasets. On average, an increase in AUC of $0.021$, pAUC of $0.031$, and a decrease in EER of $0.027$ was obtained across all the models using our proposed method on the DeeperForensics-$1.0$ dataset. Similarly, when evaluated on Celeb-DF, our proposed model obtained an increase in AUC of $0.018$, pAUC of $0.016$, and a decrease in EER of $0.021$ over the baseline CNNs in Table~\ref{Table1}. 
\emph{Thus, demonstrating the merit of our proposed approach in (a) reducing the impact of domain shift, and (b) transferability of the learned policy towards cross-dataset generalization improvement over existing CNN-based baselines}.



\subsection{Comparison with Published Results on Deepfake Detection Across Dataset}
\label{subsection53}

In Table~\ref{table3}, we compared the existing results published in~\cite{10.1007/978-3-030-58571-6_39} for deepfake detectors trained on FaceForensics++ (FF++) high quality version and tested on FaceForensics++(HQ) and Celeb-DF, with our best results obtained using EfficientNet V2-L with PPO based RL agent (see Table~\ref{table2}). 
Our proposed best model obtained \emph{equivalent performance} with most of the best-performing methods in intra-dataset evaluation. 

At the same time, our best model \emph{outperformed most of the existing methods in cross-dataset evaluation} on CelebDF with an AUC of $0.669$. Thus, obtaining state-of-the-art performance. Recall that due to the difference in the deepfake generation techniques between FaceForensics++(HQ) and Celeb-DF dataset, the performance drop of all the models in Table~\ref{table3} is significant on cross-dataset evaluation.
 
Although, Two-branch~\cite{10.1007/978-3-030-58571-6_39} obtained better results with an AUC of $0.734$ over our model in cross-dataset evaluation. This method combines representation from color and frequency domain using multi-scale Laplacian of Gaussian (LOG) operator for deep fake detection. Hence, it is not directly comparable to deep learning-based models. Further, its intra-dataset performance is quite low with an AUC of $0.931$ over our model with an AUC of $0.994$ and other existing methods in Table~\ref{table3}. 





\begin{table} []
\caption{Ablation study for choosing the optimum top-$k$ augmentations for test samples.}
\begin{center}

\scalebox{1.0}{

\begin{tabular}{l|l|c}
\hline
Augmentation Selection & FF++           & \multicolumn{1}{l}{Celeb-DF} \\ \hline
Top-$1$                  & 0.986          & 0.665                        \\
Top-$2$                  & 0.988          & 0.662                        \\
Top-$3$                  & \textbf{0.994} & \textbf{0.669}               \\
Top-$4$                  & 0.967          & 0.649                        \\
Top-$5$                  & 0.929          & 0.605                        \\ \hline
\end{tabular}}
\end{center}
\label{tab_ablation}

\end{table}

\begin{figure}
\centering
\begin{subfigure}{.3\linewidth}
    \centering
    \includegraphics[width=1.0\textwidth]{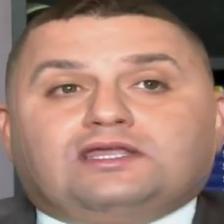}
    \caption{Original}\label{fig:image1}
\end{subfigure}
    \hfill
\begin{subfigure}{.3\linewidth}
    \centering
    \includegraphics[width=1.0\textwidth]{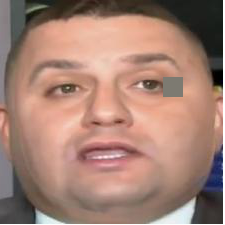}
    \caption{Top-1}\label{fig:image2}
\end{subfigure}
   \hfill
\begin{subfigure}{.3\linewidth}
    \centering
    \includegraphics[width=1.0\textwidth]{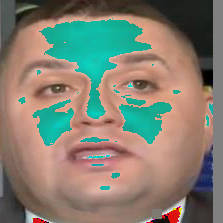}
    \caption{Top-2}\label{fig:image3}
\end{subfigure}

\begin{subfigure}{.3\linewidth}
    \centering
    \includegraphics[width=1.0\textwidth]{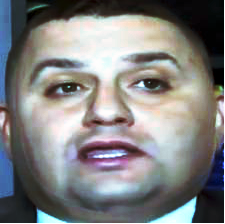}
    \caption{Top-3}\label{fig:image4}
\end{subfigure}
    \hfill
\begin{subfigure}{.3\linewidth}
    \centering
    \includegraphics[width=1.0\textwidth]{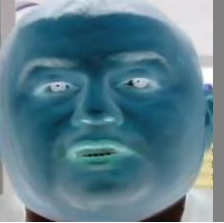}
    \caption{Top-4}\label{fig:image5}
\end{subfigure}
   \hfill
\begin{subfigure}{.3\linewidth}
    \centering
    \includegraphics[width=1.0\textwidth]{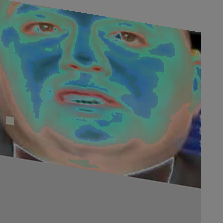}
    \caption{Top-5}\label{fig:image6}
\end{subfigure}

 \caption{The top-$k$ augmentations with $k$ ranging from $1$ to $5$ selected by PPO-based RL agent for a fake image from the FaceForensics++ dataset. These augmentations are selected from the scores obtained using Q-table generated by the RL agent. The top-$3$ augmentations obtained better results for most of the models.}
 \label{fig_aug}
\end{figure}

\section{Ablation Study: Top-$k$ Augmentations Selection}
\label{section6}
In this section, we did an ablation study to select the optimum $k$ augmentations from the list of augmentations based on the $Q(s, a)$ table that is generated by the RL agent. For the purpose of this experiment, we used the best performing EfficientNet V2-L with PPO as an RL agent trained on Faceforensics++.

Table~\ref{tab_ablation} shows the performance of EfficientNet along with PPO for top $1$ to $5$ augmentations. It can be seen that optimum results of $0.994$ AUC and $0.022$ EER are obtained for top-$3$ augmentations on FaceForensics++. 
Similarly, the highest AUC of $0.669$ and least EER of $0.362$ is obtained for top-$3$ augmentations when tested on Celeb-DF. A similar trend was observed for the DeeperForensics-$1.0$ testbed. 
Figure~\ref{fig_aug} shows the top-$5$ augmentations selected by PPO-based RL agent for a sample fake image from the FaceForensics++ dataset. 
In summary, top-$3$ augmentations obtained the optimum results for deep fake detection using our model.


\section{Conclusion and Future Work}
\label{section7}
In this paper, we improve the cross-dataset generalization of the CNN-based deep fake detectors. The proposed model applies top-$k$ augmentations to the test samples in an image-specific manner using a policy learned by an RL agent to reduce the impact of domain shift across datasets. Experimental results demonstrate the merit of systematic selection of augmentations based on the quality of each test image in cross dataset performance improvement over random augmentations. The proposed model could be used with any deep fake detection method based on supervised learning. As a part of future work, experimental evaluations will be extended with a larger number of augmentations on different datasets. The proposed RL-based method will be evaluated along with deep fake detection based on fine-grained classification~\cite{9577592} and biometric features~\cite{Dong2020IdentityDrivenDD, agarwal_protecting_2019,9717407}, for cross-dataset performance improvement.  Further, cross-comparison of our task-specific RL-based test-time augmentation approach will be conducted with automatic general-purpose data augmentation pipelines~\cite{47890}. 

\section{Acknowledgement}
The research infrastructure used in this study is supported in part
from a grant no.~$13106715$ from the Defense University
Research Instrumentation Program (DURIP) from Air Force
Office of Scientific Research.
\balance
{\small
\bibliographystyle{ieee_fullname}
\bibliography{PaperForReview}
}

\end{document}